\documentclass{llncs}
\usepackage{makeidx}  
\usepackage[colorlinks]{hyperref}
\usepackage{siunitx}
\usepackage{tabularx}
\usepackage{graphicx}
\usepackage{subcaption}

\usepackage{etoolbox} 
\makeatletter
\patchcmd{\ps@headings}{\rlap{\thepage}}{}{}{}
\patchcmd{\ps@headings}{\llap{\thepage}}{}{}{}
\makeatother
\begin{document}
%
%

%
\mainmatter              
\title{A Comparative Study of Human Activity Recognition: Motion, Tactile, and multi-modal Approaches}
\titlerunning{Hamiltonian Mechanics}  
%
\author{
Valerio Belcamino\inst{1} \and Nhat Minh Dinh Le\inst{2} \and Quan Khanh Luu\inst{3} \and \\Alessandro Carfì\inst{1} \and Van Anh Ho\inst{3}\and Fulvio Mastrogiovanni\inst{1}
}
\authorrunning{Ivar Ekeland et al.} 
%
\tocauthor{Valerio Belcamino, Dinh Minh Nhat Le, Quan Khanh Luu, Alessandro Carfì, Van Anh Ho, and Fulvio Mastrogiovanni}

\institute{TheEngineRoom, Department of Informatics Bioengineering, Robotics and System Engineering, University of Genoa, Genoa, Italy,\\
\email{valerio.belcamino@edu.unige.it},
\texttt{}
\and
Faculty of Mechanical Engineering, The University of Danang–University of Science and Technology, Danang, 54 Nguyen Luong Bang, 550000,Da Nang, Vietnam
\and
Soft Haptics Lab, School of Materials Science, Japan Advanced Institute of Science and Technology, Nomi
923-1292, Japan}

\maketitle              

\begin{abstract}
Human activity recognition (HAR) is essential for effective Human-Robot Collaboration (HRC), enabling robots to interpret and respond to human actions. This study evaluates the ability of a vision-based tactile sensor to classify 15 activities, comparing its performance to an IMU-based data glove. Additionally, we propose a multi-modal framework combining tactile and motion data to leverage their complementary strengths. We examined three approaches: motion-based classification (MBC) using IMU data, tactile-based classification (TBC) with single or dual video streams, and multi-modal classification (MMC) integrating both. Offline validation on segmented datasets assessed each configuration’s accuracy under controlled conditions, while online validation on continuous action sequences tested online performance. Results showed the multi-modal approach consistently outperformed single-modality methods, highlighting the potential of integrating tactile and motion sensing to enhance HAR systems for collaborative robotics.
\end{abstract}
\section{Introduction}\label{sec:intro}
Achieving seamless interaction between humans and robots is a key objective in human-robot collaboration (HRC) \cite{dillman}. Central to this goal is the robot’s ability to interpret and classify human actions—a task commonly referred to as human activity recognition (HAR) \cite{1570204}. Human activities are complex, encompassing interactions driven by motion \cite{sl}, those involving contact forces \cite{grasp}, and many that combine both, where movement and force are closely interconnected \cite{1570207}.

To effectively address these complexities, HAR systems in HRC require the capability to perceive both motion and tactile information. Motion-tracking approaches in the literature typically fall into two categories: wearable devices \cite{wearables} and external sensors \cite{external}. External systems often utilize optical technologies, such as RGB cameras or motion capture systems, to monitor joint movements and infer activities \cite{bodytracking}. These methods can also leverage image analysis techniques, such as optical flow \cite{opticalflow}, or enable multi-person tracking and environmental context extraction through scene segmentation \cite{multi,sceneseg}. Despite their versatility, external systems are often limited by occlusion, sensitivity to lighting conditions, and, in the case of high-end systems, prohibitive costs.

Wearable solutions, such as flex sensors \cite{flex} or inertial measurement units (IMUs) \cite{imu}, offer an alternative for motion tracking. Flex sensors are commonly embedded into garments like suits \cite{flexSuit} or gloves \cite{flexGlove} to estimate joint angles through mechanical deformation. However, they struggle with multi-axial joints and are prone to long-term wear and deformation \cite{deformation}. IMUs, by contrast, estimate spatial orientation and are often used in pairs to track relative joint motion \cite{IMUdouble}. They are well-suited for tracking joints with multiple degrees of freedom \cite{shoulder} but are affected by drift over time \cite{tokyodrift} and alignment challenges between sensor and body segment orientations \cite{sensor2segment}.

In addition to motion tracking, tactile sensing plays a pivotal role in HAR tasks involving physical interaction with objects or surfaces. Tactile sensors use different technologies such as resistive, capacitive, or piezoelectric methods \cite{pressurereview} to measure contact forces. Developers integrated these sensors into wearables like gloves, often placing the sensing elements on the fingertips or palm \cite{pressureGlove}. Vision-based tactile sensing offers another approach, capturing contact information by analyzing deformations in soft membranes under pressure. This technology provides distributed force sensing while improving safety during physical interactions due to its compliant design \cite{softtactilerev}.

The comparison of different sensor modalities for Human Activity Recognition (HAR) has been a growing focus in recent research, aiming to understand the strengths and weaknesses of each approach in various contexts. Several studies \cite{imuVcam,imuVmocap} have evaluated wearable IMUs against external vision-based systems, highlighting that IMUs achieve higher accuracy, are portable, and are resistant to occlusions and varying lighting conditions~\cite{occlusions,lighting}. However, vision-based systems remain valuable for tasks requiring full-body motion capture and multi-person tracking. Similarly, a review by Sun et al. (2022) \cite{sun2022human} extends this analysis to several other sensing modalities, including motion capture systems, infrared and depth cameras, and electromyography.
On the other hand, the tactile sensing literature for HAR is less detailed, with most studies focusing on comparing the sensors' working principles and design rather than their accuracy in HAR scenarios. Finally, in the context of multi-modal approaches, the literature predominantly emphasizes the integration of the two modalities rather than directly comparing their individual contributions.

\begin{figure}[h!]
\centering
\includegraphics[width=0.6\textwidth]{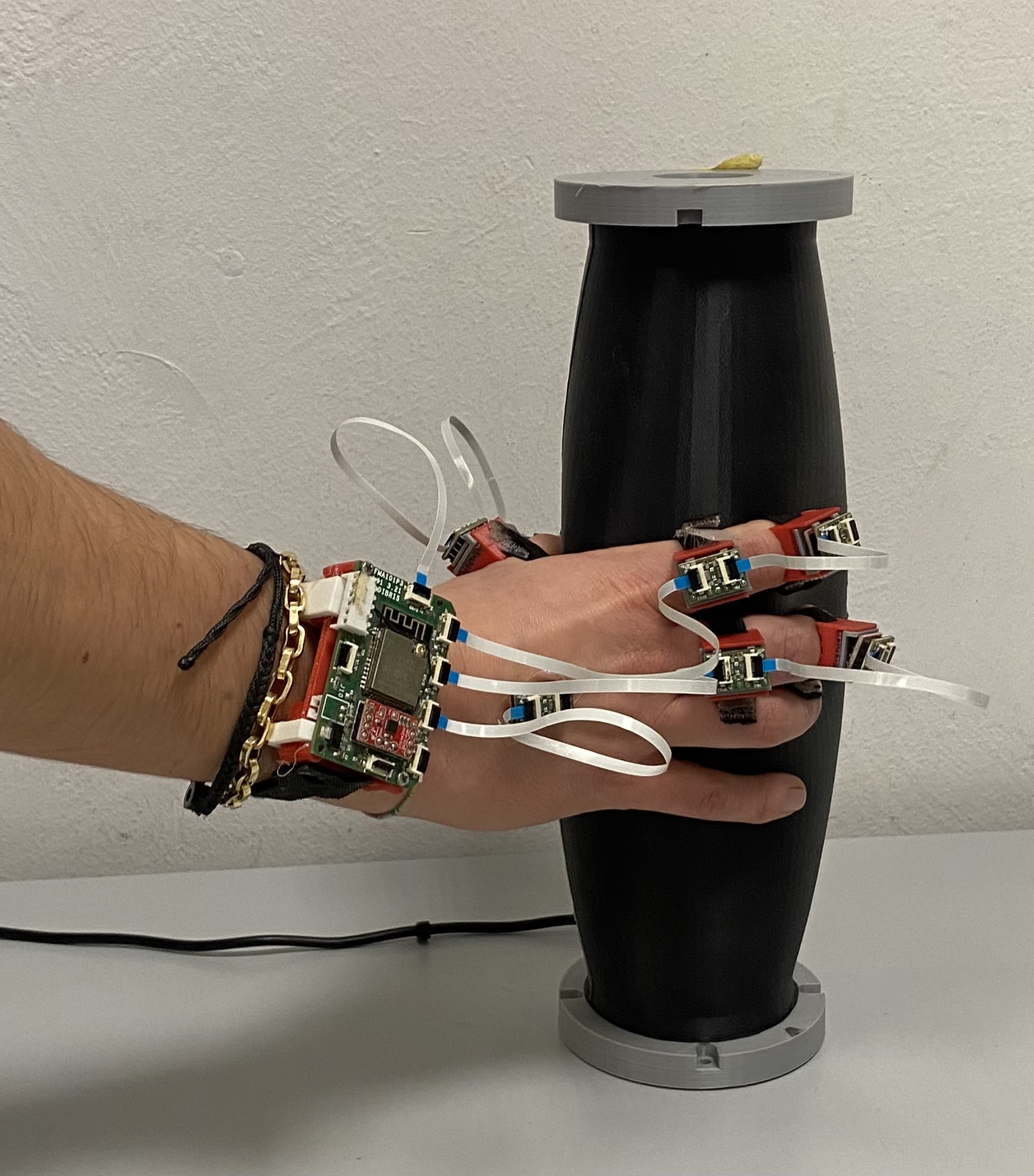}
\caption{An example of interaction with the Taclink. The user is grasping the deformable membrane while wearing the TER glove.}
\label{fig:setup}
\end{figure}

Multi-modal perception systems combining motion and tactile sensing have been explored in various domains. Applications include gait analysis \cite{s140203362}, musical performance studies \cite{piano}, human-computer interaction through gesture control \cite{hci}, and early detection of motor impairments \cite{impairmentMixed}. In robotics, multi-modal systems are frequently used in handover tasks, combining visual data from cameras or motion capture systems with tactile feedback for detecting contact forces \cite{thoduka2024multimodalhandoverfailuredetection,handoverMocap,10.1145/3656650.3656675}. However, beyond specific use cases like object handovers, the broader integration of motion and tactile sensing for activity recognition remains underexplored.

The primary objective of this work is to evaluate the capability of a vision-based tactile sensor to classify human activities accurately. The classification performance of this sensor is influenced by factors such as the resolution and frame rate of its embedded cameras, as well as the mechanical properties of its deformable membrane. To contextualize these results, we compared its performance to that of IMU-based motion data, which is one of the most widely explored approaches in the literature for human activity recognition (HAR). We selected IMUs also for their ability to capture the complex kinematics of joints with multiple degrees of freedom and their robustness against occlusions. Finally, we propose a multi-modal approach that integrates both sensing modalities, where the tactile data complements the motion data. This combined method aims to enhance recognition accuracy by leveraging the complementary strengths of motion and contact force sensing.
Specifically, we evaluate three configurations: (i) motion data using IMU sensors, (ii) tactile data using a vision-based tactile sensor, and (iii) a multi-modal approach that integrates both modalities. 
To assess the performance of these configurations, we classify 15 hand gestures commonly used in tactile sensing research \cite{touchactions2} and social interaction studies \cite{socialtouch}. These gestures represent diverse characteristics such as contact area, pressure levels, and frequency of interaction. Finally, our evaluation includes both tests in static conditions on a segmented dataset to quantify classification accuracy, and tests on continuous sequences to assess online performance.

\section{Methodology}
\label{sec:methodology}
As we previously introduced, this work aims to compare the effect of different sensorial features on the classification of human gestures. To achieve this goal, we collected data from tactile and motion sensors during the execution of 15 of the most common gestures in social interaction and the benchmarking of tactile sensors~\cite{touchactions}. Then, we trained multiple deep learning models with different input configurations and compared their results.

\begin{table}[!t]
\begin{minipage}{0.68\textwidth} 
\begin{tabular}{l@{\hspace{0.5cm}}l@{\hspace{0.5cm}}l} 
\textbf{Acronym} & \textbf{Configuration Name}  & \textbf{IMUs} \\ 
\hline
8FG  & Full Glove             & ABCDEFGH \\ 
4CPW & Core Phalanges and Wrist  & BDFH \\
4CPB & Core Phalanges and Back & BDFG \\
3CP  & Core Phalanges         & BDF \\
2TI  & Thumb and Index        & BD \\
3TIB & Thumb, Index and Back  & BDG \\
2WB  & Wrist and Back         & GH \\
\end{tabular}
\end{minipage}
\hfill 
\begin{minipage}{0.28\textwidth} 
\includegraphics[trim={0 0 0 2},clip, width=\textwidth]{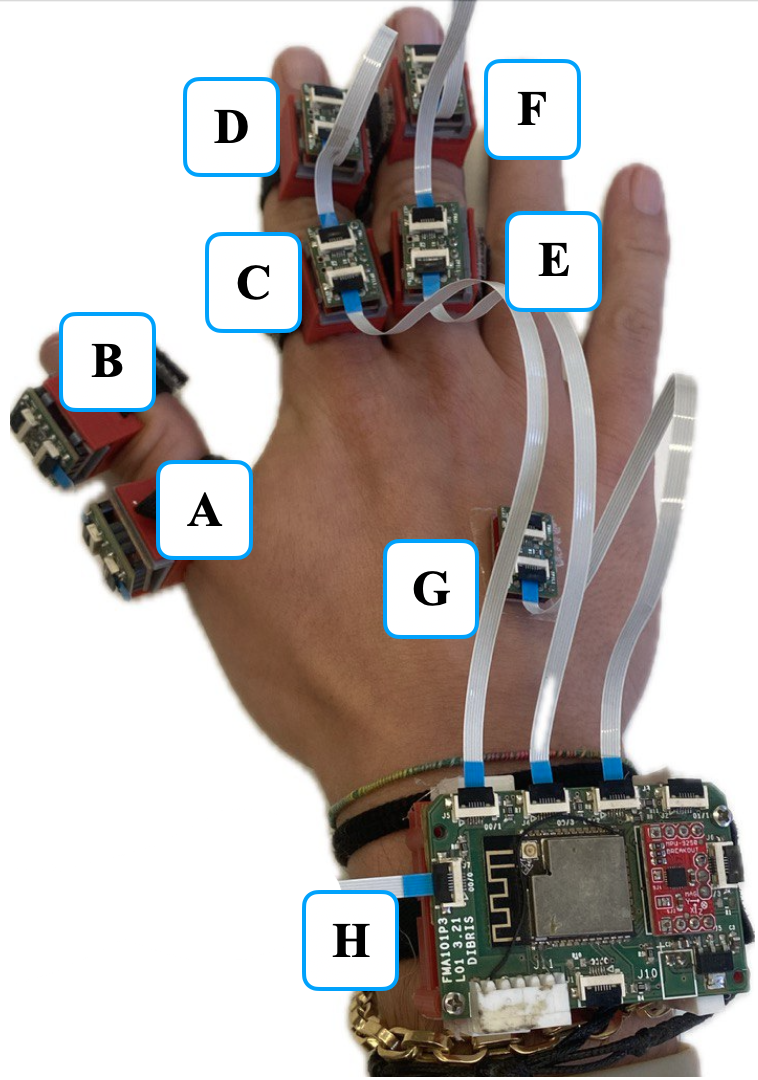} 
\end{minipage}%
\vspace{0.2cm}
\caption{The left table lists the seven data glove configurations used in this work with their acronym, full name, and reference to the included sensors. The image on the left shows a hand wearing the TER glove in the full glove configuration and the sensors' labels.}
\label{fig:glove_configurations}
\end{table}

Our comparison comprises three categories referring to the different sensing modalities used as input: tactile-based classification (TBC), motion-based classification (MBC) and multi-modal classification (MMC). 

Tactile-based classification (TBC) utilizes input from the Taclink \cite{taclink1}, a vision-based tactile sensor designed to sense large-scale contact forces. The Taclink features a cylindrical body made of silicon polymer, measuring \SI{23}{\cm} in height and \SI{4}{\cm} in radius, with an active sensing area of approximately \SI{578.05}{\cm^2}. Reflective markers, arranged in a uniform grid, cover the Taclink inner surface, while two RGB cameras are positioned at the top and bottom of the cylinder. These cameras operate at 30 frames per second with a resolution of 1080p, capturing marker movements to estimate surface deformations. Its design allows the Taclink to measure the magnitude of contact forces and the contact area. While it is designed to be mounted as an additional link on a manipulator, in this study, we fixed the Taclink on a table.

Motion-based classification (MBC) leverages data from the TER glove \cite{terglove}, a modular wearable device equipped with 9-axis IMU sensors. Each module measures 3-axis accelerations, angular velocities, and magnetic field values, streaming the data to a workstation via Wi-Fi. TER glove modular architecture allows for up to 12 sensors; however, increasing their number impacts both sampling frequency and battery life. For this study, we selected a configuration with 8 IMUs to balance performance and autonomy, achieving a sampling rate of \SI{40}{\hertz} with approximately one hour of battery life. To match the lower frequency of the cameras, we downsampled the sequences collected with the glove to \SI{30}{\hertz} using the timestamps collected in for the TBC. Sensors were positioned on the thumb, index, and middle fingers (two sensors each), as well as on the back of the hand and wrist. This arrangement was chosen to capture the intricate kinematics of the hand, particularly considering the coupling between the middle, ring, and little fingers in our action set.

Multi-modal classification (MMC) combines TER glove and Taclink to exploit their complementary capabilities. Motion data from the IMU sensors captures detailed hand kinematics, while tactile data from the Taclink provides insights into contact force magnitude and area. By integrating these two modalities, MMC aims to enhance classification accuracy by analyzing motion and tactile interactions simultaneously, offering a richer understanding of human activities.

For the TBC, we conducted tests under three distinct conditions: using as input features the videos from the top camera (TBC-1top), from the bottom camera (TBC-1bot) separately, and then from both (TBC-2). 
In the case of the MBC, we conducted tests with seven different sensor configurations, varying the number of IMU considered in the classification. The various sensor configuration, shown in Fig. \ref{fig:glove_configurations}, are: i) full glove (8FG), ii) core phalanges and wrist (4CPW), iii) core phalanges and back (4CPB), iv) core phalanges (3CP), v) thumb and index (2TI), vi) thumb, index and back (3TIB), vii) wrist and back (2WB). For a clearer comparison, the acronym for each configuration starts with a number, which refers to the number of sensors considered. 8FG coincides with the physical setup of our data glove, as shown in Fig. \ref{fig:glove_configurations}, with one IMU on the intermediate and proximal phalanges of the index and middle fingers, 2 IMUs on the proximal and distal phalanges of the thumb, one on the wrist and one on the back of the hand. 3CP has one sensor on the intermediate phalanges of the index and middle fingers and one on the distal phalange of the thumb, whereas 4CPB also tracks the back of the hand. 3PIT is the same as 4CPB without the tracking of the middle finger. Additionally, 2TI includes one sensor on the distal phalange of the thumb and one on the intermediate of the index, with 3TIB having an additional IMU on the hand's back. The last configuration is 2WB, which only comprises the back and wrist modules. It is important to notice that, since the operating frequency and battery life of the glove are influenced by the number of active sensors, all 8 physical sensors have been connected during the recording process and the experiments with the different configurations have been carried out offline discarding the unnecessary features.
Lastly, for the MMC, we employed motion and tactile sensation, relying on both video streams and the same seven IMU configuration described for the MBC. 

\begin{figure}[!t]
\centering
\begin{subfigure}[t]{0.45\textwidth} 
    \centering
    \includegraphics[width=\textwidth]{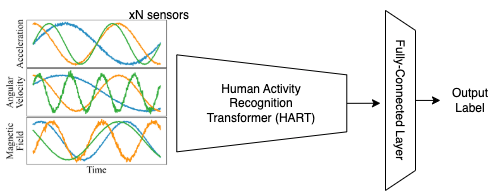} 
    \label{fig:image1}
\end{subfigure}
\hfill
\begin{subfigure}[t]{0.45\textwidth}
    \centering
    \includegraphics[width=\textwidth]{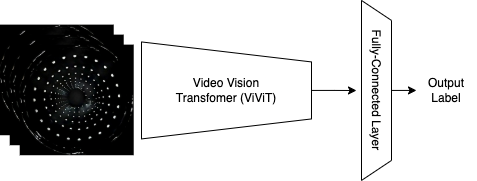} 
    \label{fig:image2}
\end{subfigure}
\caption{On the left is the IMU-based model, whose input size depends on the sensor configuration. On the right is the tactile-based model, which relies on a single video input.}
\label{fig:pair1}
\end{figure}

\begin{figure}[!t]
\centering
\begin{subfigure}[c]{0.45\textwidth} 
    \centering
    \includegraphics[width=\textwidth]{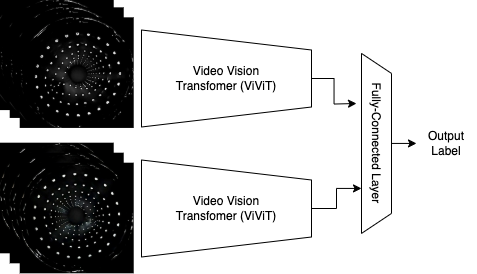} 
    \label{fig:image3}
\end{subfigure}
\hfill
\begin{subfigure}[c]{0.45\textwidth} 
    \centering
    \includegraphics[width=\textwidth]{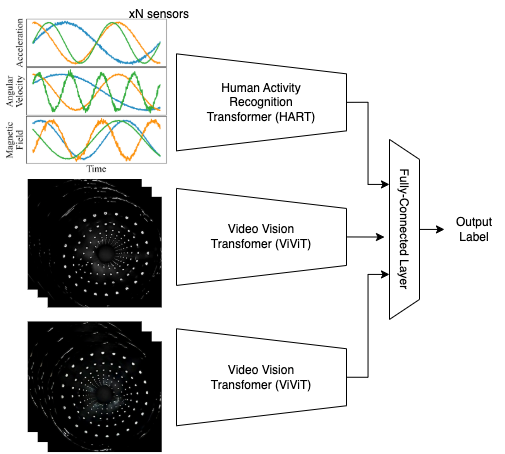} 
    \label{fig:image4}
\end{subfigure}
\caption{On the left is the tactile-based approach, which relies on both cameras. On the right is the multi-modal solution combining tactile and motion information.}
\label{fig:pair2}
\end{figure}

In this work, we employed multiple neural network architectures tailored to the specific input data for each classification approach. Motion-based classification (MBC) relies on a Human Activity Recognition Transformer (HART) architecture \cite{s22051911}, which is well-suited for processing temporal sequences. The HART model extracts motion-related features such as heading, velocity profiles, and inter-joint synergies from the TER glove’s IMU data, enabling a detailed analysis of dynamic hand movements. For different configurations of the glove (e.g., varying the number of IMU modules), we scaled the input size of the HART model to accommodate the corresponding number of features.
Tactile-based classification (TBC) utilizes the ViViT architecture \cite{arnab2021vivitvideovisiontransformer}, designed for analyzing spatial-temporal patterns in video data. When a single video input is needed (i.e., TBC-1top or TBC-1bot),  one ViViT backbone processes the video stream, extracting features such as contact shapes, pressures, and deformations. Instead, when both RGB streams from the Taclink are in use (i.e., TBC-2), two parallel ViViT backbones process their respective video stream independently before their outputs are stacked and passed through a fully connected layer for final classification. 
Multi-modal classification (MMC) integrates motion and tactile data using a combination of HART and ViViT architectures. Specifically, this configuration employs two ViViT backbones to process the dual video streams from the Taclink and one HART branch to process IMU data from the TER glove. During the late fusion step, the features extracted by all three branches are combined into a fully connected layer, enabling the model to classify  based on a comprehensive understanding of both motion and contact forces.
We adopted late fusion to combine multiple data sources for its flexibility in processing heterogeneous data types independently through specialized branches \cite{latefusion}. This mechanism allows each modality to contribute high-level features before merging \cite{latefusion2}, helping the network to focus on modality-specific patterns—such as distinct motion cues from IMUs or tactile features from video streams—before integrating them at the decision stage.
This design ensures that spatial-temporal patterns unique to each sensor type are leveraged effectively, enhancing classification accuracy for complex activities involving motion and contact.
OOur approach, illustrated in Figs. \ref{fig:pair1} and \ref{fig:pair2}, extends existing frameworks, combining the features extracted from ViViT \cite{arnab2021vivitvideovisiontransformer} and HART \cite{s22051911} in a multi-branch setup. The flexibility of this design allows seamless adaptation to single-modality inputs (e.g., MBC or TBC) or multi-modality strategies (e.g., MMC), ensuring robust performance across diverse configurations while enabling fair comparison between different sensing solutions.

We tested all possible input configurations using the described models on manually segmented data. These tests allowed us to compare the different sensing modalities and identify the best input configuration for each of the three sensing modalities. Following this, we conducted additional tests to evaluate the online classification capabilities of our models, comparing the optimal sensing configurations identified in the first phase. We performed these tests on continuous sequences of human actions to determine the best sensing modality for the given scenario.

\section{Experimental Setup}
\label{Experimental Setup}

We structured the experimental setup to test the capabilities  of proposed approaches to classify  15 human actions. These actions encompass a wide range of tactile and motion characteristics, including different areas and intensities of contact and frequency of interaction. 
Table \ref{tab:action_characteristics} lists the considered actions, providing brief descriptions and an overview of their contact properties.
A video showcasing examples for each action is available\footnote{\scriptsize{\href{https://youtu.be/GKYB5cKDM04}{youtube.com/watch?v=GKYB5cKDM04}}}.
Our experimental setup comprises the TER glove and Taclink sensors, a Linux workstation for data collection and the KAGAYAKI HPC server\footnote{\scriptsize{\href{https://www.jaist.ac.jp/iscenter/en/mpc/kagayaki/}{jaist.ac.jp/iscenter/en/mpc/kagayaki/}}}, used for the training the models.

\begin{table}[t]
\renewcommand{\arraystretch}{1} 
    \centering
    \scriptsize
    \begin{tabular}{lllll}
        \textbf{Action} & \textbf{Area} & \textbf{Pressure} & \textbf{Frequency} & \textbf{Description} \\
        \hline \\[-2ex]
        Pinching    & Small  & High   & Medium    & Using two fingers to press against a single point. \\
        Pulling     & Large & High & Low    & Gripping and slightly tugging at the surface. \\
        Pushing     & Large  & High   & Low    & Applying steady force with the hand or fingers. \\
        Rubbing     & Medium & Medium    & High   & Moving fingers back and forth repetitively. \\
        Patting     & Large  & Medium    & High & Intermittent contact with open palm. \\
        Tapping     & Medium  & Medium & High   & Quick, small contact points with the fingertips. \\
        Scratching  & Small  & Low    & High   & Dragging fingertips with minimal pressure. \\
        Lingering   & Medium  & Low    & Low    & Holding a steady contact without movement. \\
        Massaging   & Large  & Medium & Medium & Applying circular pressure with a broad surface. \\
        Squeezing   & Large  & High   & Low    & Compressing the surface with multiple fingers. \\
        Trembling   & Small  & Low    & Medium   & Slight oscillations exerted by the fingertips. \\
        Shaking     & Large  & Medium & High   & Applying larger, rapid oscillations or vibrations. \\
        Stroking    & Medium & Low    & Medium    & Smoothly moving the palm up and down. \\
        Poking      & Small  & Medium & Medium    & Applying localized force with a single finger. \\
        Idle      & N/A  & N/A & N/A    & Includes actions with no physical interaction. \\
    \end{tabular}
    \vspace{0.5cm}
    \caption{The list of actions considered in our study. The first column shows the names of the actions, while the second one holds a discrete value that describes the extent of the contact surface for each action, ranging from Small to Large. The second and third columns show another discrete value that describes the amount of pressure and the frequency for each action. Lastly, there is a textual description of the action. For the \textit{idle} action, columns 2,3, and 4 do not have a value since there should be no contact while it happens.}
    \label{tab:action_characteristics}
\end{table}

In our study, we considered two experimental scenarios: offline and online.  The offline scenario aimed to collect data for training and evaluate the system's ability to classify pre-segmented actions. For this phase, we used the TER glove in the 8FG configuration to ensure a consistent sampling frequency across all trials. Participants performed each of the 15 actions individually while interacting with the Taclink sensor, creating a segmented dataset. The recording for each action lasted 1 minute, during which the participant repeated the action multiple times with their dominant hand. Participants repeated the process with their non-dominant hands as well. This procedure resulted in 30 minutes of data per participant, and with 8 participants contributing to the dataset, we collected approximately 4 hours of recorded data.
To prepare the tactile data for processing, we cropped the frames to a square ratio, removing the lateral black borders to focus on the circular marker pattern visible in the cameras, as shown in Figure \ref{fig:pair2}. We then converted the images to greyscale and resized them to a resolution of 224x224 pixels, chosen to balance computational efficiency and image quality. Each 1-minute recording was divided into 3-second sequences, producing over 4,000 samples. We selected this sequence length as it is sufficient to capture even the longest actions. However, as segmentation was performed automatically, some sequences may contain partial actions or no action at all if idle periods fell within the window. The video frames were standardized by calculating the mean and standard deviation within each dataset split, while IMU data from the TER glove was normalized using its full-scale sensor range. The resulting dataset was divided into three partitions: 60\% for training, 20\% for validation, and 20\% for testing.
The training phase was conducted multiple times for each model and each sensor configuration introduced in the previous section. Tactile-based classification (TBC) involved 3 models: one for each camera individually and one combining both cameras. For motion-based classification (MBC), we trained 7 models corresponding to the different configurations of the glove. Finally, multi-modal classification (MMC) consisted of 7 models, integrating both tactile and motion data for each glove configuration. For the MMC both cameras were used in each trial.

In the second scenario, we assessed the ability of the models to perform online classification on continuous sequences of human interactions. Using the experimental setup previously introduced, we gathered a new dataset. This new dataset comprises longer unsegmented sequences composed of multiple actions. For this, we recruited 4 participants who performed all 15 actions in a continuous sequence interspersed by brief pauses. To ensure a balanced dataset, we predefined the order of actions for each participant. Each participant completed the sequence twice, once with each hand, resulting in trials lasting approximately 15 minutes per participant and a total dataset duration of about 2 hours. Notably, two participants from this test also participated in the earlier experiment.
To ensure accurate labeling, an experimenter manually annotated the sequences. The annotation relied on the predetermined action order and video recordings from Taclink’s cameras, which allowed us to identify the start and end points of each action. We labelled pauses between actions as the \textit{idle} class.

For the online classification testing, we focused exclusively on the best-performing models identified for each input modality during prior evaluations. Specifically, these were TBC-2 for the tactile data, MBC-8FG for the IMU data, and MMC-8FG for multi-modal inputs. These models were selected based on their performance in the offline classification, ensuring that only the most accurate solutions were employed  here.

We processed each sequence in new dataset using the neural network models trained on segmented data and collected the predicted labels for each frame. We employed a moving window approach for processing the temporal sequences, with a buffer dynamically updated with incoming sensor data. We set the window size to 90 samples, corresponding to the 3-second segments used in the offline dataset.
The preprocessing previously described were also applied here at every timestep. These included image transformations such as cropping, scaling, grayscale conversion, and standardization, as well as normalization of IMU data.
To assess the models performances in online classification, we compared their output labels to the ground truth on a frame-by-frame basis.

\section{Results}
\label{Results}
Here, we present the results of our comparative analysis, evaluating the performance of tactile-based classification (TBC), motion-based classification (MBC), and multi-modal classification (MMC). As previously described, we tested each classification approach under different configurations and evaluated them using the $F_1$ score. Additionally, online testing was conducted using the most accurate configuration for each sensor modality.

\begin{figure}[h!]
\centering
\includegraphics[width=\textwidth]{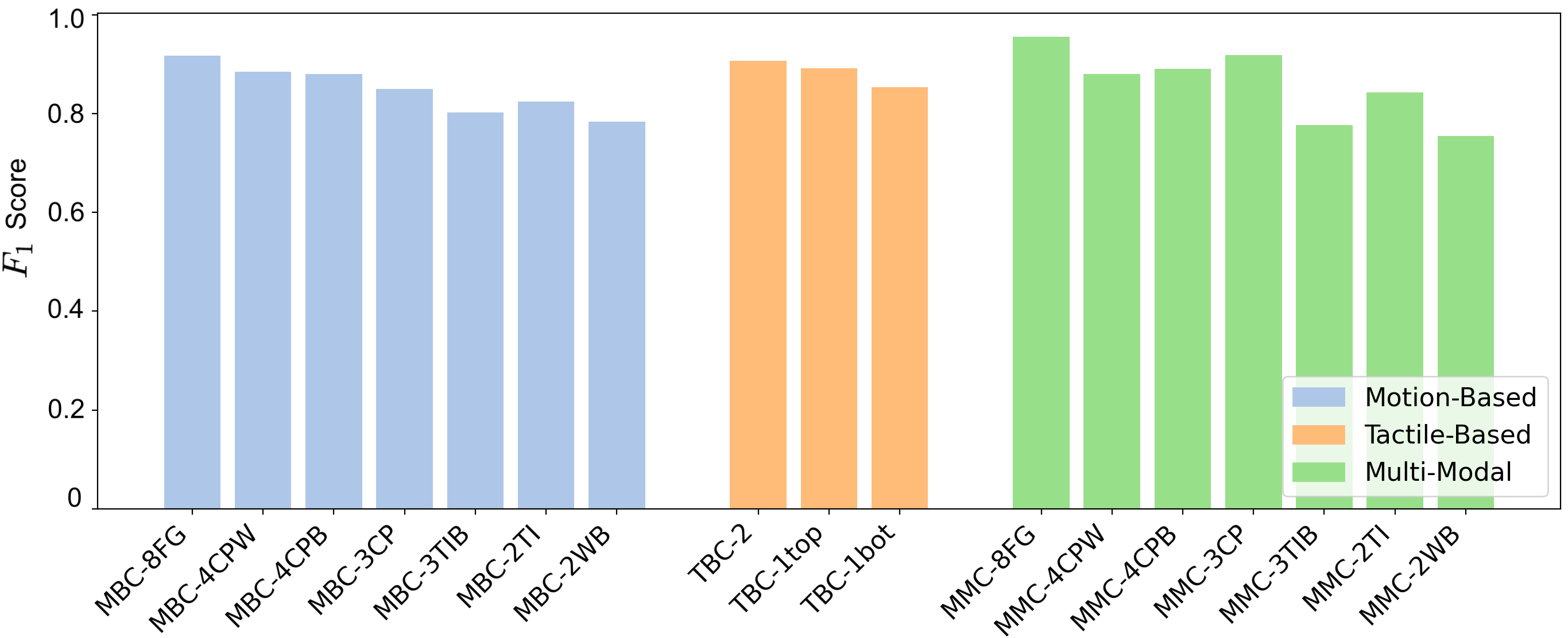}
\caption{The plot shows the offline classification results for all the models considered in our comparison. The blue bars refer to the motion-based models, and the orange ones to the models that rely on tactile perception. Lastly, the multi-modal models are shown in green.}
\label{fig:classification_results}
\end{figure}

In the offline scenario, we tested all configurations using a segmented dataset, and the results are shown in Fig. \ref{fig:classification_results}. 
Among tactile-based models, TBC-2 demonstrated the best performance, with an $F_1$ score of 90.66\%. This result underscores the advantage of integrating tactile information from multiple perspectives to capture more accurate contact features. In contrast, TBC-1top achieved 89.14\%, while TBC-1bot showed slightly lower performance at 85.37\%. The disparity between the top and bottom camera configurations can be attributed to the fixed position of the Taclink sensor during data set recording. Since the sensor was mounted on a tabletop and never flipped, the top camera had a more favorable angle, enabling it to extract more features from the images.
For the MBC approach,the full glove (MBC-8FG) achieved the best performance, with an $F_1$ score of 91.74\%. Other configurations demonstrated lower performance, with performances decreasing as fewer IMUs were used. Specifically, MBC-4CPW achieved 88.47\%, MBC-4CPB reached 87.99\%, and MBC-3CP obtained 84.97\% $F_1$ score. The lowest-performing configurations were MBC-2TI (82.40\%), MBC-3TIB (80.28\%), and MBC-2WB (78.37\%). Surprisingly, we achieved performances close to the full setup using only half the sensors, specifically by focusing on the intermediate phalanges. These links demonstrated a higher range of motion in the set of actions we selected, making them particularly effective for capturing the dynamics of these actions. These findings suggest that careful selection of sensor placement, informed by the specific motion patterns of the target actions, can significantly reduce hardware complexity without sacrificing much performance. As expected, the IMUs demonstrated excellent performance for this task. Notably, even with the minimal configuration of just two IMUs placed on the back and wrist (MBC-2WB), we achieved respectable results, with an $F_1$ score over 78\%.
The MMC approach, which integrates motion data from the TER glove and tactile data from the Taclink, demonstrated the highest overall performance in the offline tests. With the complete configuration (MMC-8FG), this approach achieved an accuracy of 95.60\%, outperforming both MBC and TBC individually. This result highlights the complementary nature of motion and tactile data, with each modality compensating for limitations in the other. For example, tapping and patting have similar contact features that are difficult to distinguish using Taclink images alone. However, these gestures exhibit significantly different joint states, which IMUs can capture effectively, enabling easier and more accurate classification. Other multi-modal configurations also performed well: MMC-3CP achieved 91.82\%, MMC-4CPB reached 89.07\%, MMC-3PIT obtained 87.98\%, and MMC-2TI achieved 84.32\%. However, configurations with reduced sensor coverage, such as MMC-3TIB (77.70\%) and MMC-2WB (75.50\%), showed significantly lower performance, emphasizing the need for comprehensive sensor coverage when combining modalities.

Regarding inference time, the MBC approach was the fastest among the tested models. The TBC with a single video input required approximately 6.1 times more time compared to MBC. When using dual video input, the TBC inference time increased to 12.2 times that of MBC. The MMC approach, integrating both motion and tactile data, required the highest inference time, taking 12.51 times longer than the MBC. These results underscore the trade-off between accuracy and computational efficiency when combining modalities. Nonetheless, each model could run faster than 30Hz on our workstation without causing any issues with our data streaming pipeline. This phase of online classification was performed on a laptop equipped with an Intel i5 7th gen processor and an Nvidia GeForce GTX 1070 Ti.

In online testing, we evaluated the three approaches using unsegmented sequences of human activities. For this phase, only the most accurate configurations were considered: both cameras in the Taclink sensor (TBC-2) and all eight IMUs in the TER glove (MBC-8FG). The MMC model achieved a frame-level accuracy of 83.92\%, outperforming both MBC (79.54\%) and TBC (71.03\%). 
The overall decrease in performance compared to the offline test can be attributed to the increased complexity of the task and the use of a moving window approach for online classification. This approach introduces a small delay in classification, resulting in a misalignment between the ground truth and classified labels on a frame-by-frame basis. Despite this challenge, MMC and MBC delivered good results across the 15 classes.
For TBC, however, the performance drop was more pronounced, likely due to the deformation of the Taclink sensor during interactions. The soft membrane of the sensor takes time to return to its initial state after a user releases it, making idle actions particularly difficult to classify in this setup. In contrast, the MMC approach mitigated this issue by integrating motion data, which helped the model distinguish actions more effectively, even in cases where the tactile data was temporarily unreliable.

\section{Conclusions}

Our study evaluates and compares the performance of motion-based, tactile-based, and multi-modal approaches for human activity recognition. By leveraging data from IMU-equipped gloves and vision-based tactile sensors, we systematically assessed these modalities under both offline and online conditions. Our findings demonstrate that the multi-modal approach consistently outperforms single-modality methods, achieving an $F_1$ score of 95.60\% during offline validation and maintaining robust performance in online testing with a frame-level accuracy of 83.92\%.

Offline testing highlighted the significant role of sensor placement and coverage in motion-based classification, as configurations with more IMUs and strategic sensor positioning demonstrated higher performances. Similarly, the tactile-based approach benefited from combining video streams from multiple cameras, emphasizing the importance of multiple points of view in vision-based tactile perception. Notably, integrating these two sensing modalities in a multi-modal framework proved to be particularly effective, leveraging the complementary strengths of motion tracking and contact force perception.

In online testing, the performance of the multi-modal approach remained strong, with the MMC model achieving a frame-level accuracy of 83.92\%. 
Even in the online classification, the multi-modal approach outperformed both the tactile and the motion-based. In this task, we observed a general performance drop, which can be attributed to the online classification delay, and it was particularly noticeable in the TBC due to the sensor deformation. Interestingly, the multi-modal approach did not suffer from the same problem, as the motion data effectively compensated for the deformation, providing more reliable results even in challenging scenarios.

In summary, this work highlights the value of combining motion and tactile data for human activity recognition, particularly in contexts involving online computations. The contributions of this study include the evaluation of different sensor configurations and the demonstration of how multi-modal fusion can enhance classification accuracy. 

\section*{Acknowledgement}
 This research is partially supported by the Italian government under the National Recovery and Resilience Plan (NRRP), Mission 4, Component 2, Investment 1.5, funded by the European Union NextGenerationEU programme, and awarded by the Italian Ministry of University and Research, project RAISE, grant agreement no. J33C22001220001.



\bibliographystyle{IEEEtran}

\bibliography{bibliography}

\end{document}